\documentclass{article}
\usepackage{spconf,amsmath,graphicx,amssymb}
\usepackage{algorithm}
\usepackage{algorithmic}
\ninept

\def\v#1{\mathbf{#1}}
\def\L#1{\mathcal{#1}}

\title{Few-shot acoustic event detection via Meta Learning}
\name{Bowen Shi\textsuperscript{1*}\thanks{*Work done at Amazon}, Ming Sun\textsuperscript{2}, Krishna C. Puvvada\textsuperscript{2}, Chieh-Chi Kao\textsuperscript{2}, Spyros Matsoukas\textsuperscript{2}, Chao Wang\textsuperscript{2}}
\address{\textsuperscript{1}Toyota Technological Institute at Chicago\\
\textsuperscript{2}Amazon\\
bshi@ttic.edu, \{mingsun,puvvav,chiehchi,matsouka,wngcha\}@amazon.com}

\begin{document}
%
\maketitle
\begin{abstract}
  We study few-shot acoustic event detection (AED) in this paper. Few-shot learning enables detection of new events with very limited labeled data. Compared to other research areas like computer vision, few-shot learning for audio recognition has been under-studied. We formulate few-shot AED problem and explore different ways of utilizing traditional supervised methods for this setting as well as a variety of meta-learning approaches, which are conventionally used to solve few-shot classification problem. Compared to supervised baselines, meta-learning models achieve superior performance, thus showing its effectiveness on generalization to new audio events. Our analysis including impact of initialization and domain discrepancy further validate the advantage of meta-learning approaches in few-shot AED.
\end{abstract}
\begin{keywords}
Acoustic event detection, few-shot learning, meta learning
\end{keywords}
\section{Introduction}
\label{sec:intro}
Acoustic event detection (AED) is the task of detecting whether certain events occur in an audio clip. It can be applied in many areas such as surveillance \cite{surveil_1, surveil_2}, and recommendation systems \cite{recommendation}. 
Current state-of-the-art AED models are data-hungry and large number of labeled data is needed to achieve high performance on detecting target event \cite{rcrnn}, which causes problem for detecting events where labels are limited. One potential solution is semi-supervised learning which incorporates unlabeled data for model training. Multiple methods belonging to this category \cite{tri_train, Lu2018, Lin2019} have been explored for AED. However those approaches fail to generalize to few-shot scenarios where only very few samples (e.g., $<10$) are available for target event. Few-shot learning for AED is useful in practice not only because there exists large variety of audio events and labels of rare events are extremely limited. Besides, some audio events like doorbell sounds can be different for different households and there is no universal definition for such event. Efficient few-shot learning facilitates the personalization of AED in real life.

Few-shot learning tasks have been increasingly studied in literature and often rely on meta-learning approaches including MAML (Model-Agnostic Meta-Learning)\cite{maml} and Prototypical networks \cite{proto}. Most such works are done in computer vision \cite{mini_imagenet,Koch2015SiameseNN} or natural language recognition\cite{Yu2018metric} while very little work has been done in audio-related tasks. \cite{att_chou} is the only existing work on sound recognition to our knowledge. However it studies multi-way classification while we focus on multi-label classification (detection) here. To our knowledge few-shot detection problem has not been thoroughly studied before, though it bears similarity with multi-way few-shot classification. Formulation of this problem will be discussed in detail in section \ref{sec:method}. Besides we will also study several typical meta-learning approaches under few-shot AED setting and compare them with traditional supervised models in different scenarios (e.g., domain shift).

\section{Methods}
\label{sec:method}

\subsection{Basic AED model}

For AED, given feature of audio signal $\v x$ (e.g. Log-mel Filterbank Energy) the task is to train a model $f$ to predict a multi-hot vector
 $\v y\in\{0, 1\}^C$ with $C$ being the size of event set 
and $y_c$ being a binary indicator whether event $c$ is present in $\v x$. It is a multi-label classification problem and prediction $f(\v x)$ is not
a distribution over event set since multiple events can occur in $\v x$.
In supervised setting, we train model $f$ using cross-entropy loss (see equation \ref{eq:sup_loss}), where $w_c$ is
the penalty on mis-classifying true positive samples of class $c$. $w_c$ serves the purpose of balancing losses between positive and negative instances and is tuned as hyper-parameter in practice.

\begin{equation}
\footnotesize
  \centering
\label{eq:sup_loss}
L(f,D)=-\displaystyle\sum_{(\v x, \v y)\in D}\displaystyle\sum_{c=1}^C\{w_c y_c\log f^{c}(\v x)+(1-y_c)\log(1-f^c(\v x))\}
\end{equation}

\subsection{Few-shot AED setup}

We aim to obtain a model that can be trained to perform good detection given few labeled examples for new event. Suppose we are given two non-overlapping sets of classes (events) $\mathcal{C}_{train}$ and $\mathcal{C}_{test}$.
The task is to train a detection model $f$ using labeled data of classes from $\mathcal{C}_{train}$ and to evaluate on data of classes in $\mathcal{C}_{test}$.

Few-shot learning problem is commonly tackled in a meta-learning setting, which is also called episodic setting in literature \cite{vinyals2016}. It operates on meta datasets where each element is a dataset and thus instantiates a task, which is multi-label classification here. For two tasks, events (classes) to detect can differ. Models are trained with \emph{meta-training set} and will be tested on \emph{meta-test set}. There is \emph{no class overlap} in the two sets. Similar to supervised learning, there often exists a \emph{meta validation set} which is used for hyper-parameter tuning and model selection. This special setup allows model to generalize across classes since it is trained with a set of tasks while each consists of different classes.

Now we formally define the setup. A dataset $\L D$ is a set of $<$audio, events$>$ pairs. A meta-set $\L M$ is a set of datasets. Each dataset $\L D$ instantiates a $K$-way binary classification problem where $K$ is number of events. $\L D$ is divided into two parts, ``training'' subset $\L S$ and ``test'' subset $\L Q$. To avoid confusion with meta-training and meta-test set mentioned above, $\L S$ and $\L Q$ are herein called support and query set respectively. Under this setting, model $f$ is trained with meta-training set $\L M_{train}$ and evaluated with meta-test set $\L M_{test}$. More concretely, $\forall \L D_i\in \L M_{test}$, model $f$ is required to make prediction on query set $\L Q_i$ (no label) given its support set $\L S_i$ (with label). We also define \emph{shots} $N$ as the number of positive samples in support set $\L S$. A typical value of $N$ would be 1 $\sim$ 5 in few-shot setting. Note we only constrain number of positive samples per event, which is different from multi-way few-shot classification problems where $N$ refers to number of labeled data for each class. This is for practical considerations since negative samples for one particular event can be much easier to acquire. From a typical labeled dataset (e.g., AudioSet), meta datasets can be constructed through sampling. Details on the construction process is shown in alogithm \ref{alg:meta-const}.


\begin{algorithm}[tb]
  \caption{Construction of training, validation and test sets for $K$-way $N-$shot detection. $N^{-}_S$, $N^{-}_Q$ are number of negative samples for support and query set respectively. All sampling is without replacement.
 $\textsc{Sample}(X, N, C)$ denotes a set of $N$ elements of event set $C$ chosen uniformly at random from set $X$}
    \label{alg:meta-const}
    \begin{algorithmic}
      \REQUIRE Dataset $\L D=\{(\v x_1, y_1), ..., (\v x_n, y_n)\}$. Whole set of events $\L C$ (label set).
      \ENSURE Training/validation/test set $\L M_{train}$/ $\L M_{val}$/ $\L M_{test}$
      \STATE Split $\L C$ into three disjoint sets $\L C_{train}$, $\L C_{val}$, $\L C_{test}$
      \STATE Split $\L D$ into $\L D_{train}$, $\L D_{val}$, $\L D_{test}$ according to $\L C_{train}$, $\L C_{val}$, $\L C_{test}$
      \FOR{$x$ in $\{train, val, test\}$}
      \FOR{$t$ in $\{1,...,|\L M_x|\}$}
      \STATE Randomly sample $K$ events $\L C_{t}=\{c_1, c_2,..., c_K\}$ from $\L C_x$
      \FOR{$k$ in $\{1,2,...K\}$}
      \STATE $\L S_t \gets \L S_t \cup \textsc{Sample}(\L D_x, N, \{c_k\})$
      \STATE $\L Q_t \gets \L Q_t \cup \textsc{Sample}(\L D_x, Q, \{c_k\})$
      \ENDFOR
      \STATE $\L S_t \gets \L S_t \cup \textsc{Sample}(\L D_x, N^{-}_S, \L C_x \setminus\L C_t)$
      \STATE $\L Q_t \gets \L Q_t \cup \textsc{Sample}(\L D_x, N^{-}_Q, \L C_x \setminus\L C_t)$
      \STATE $\L M_x \gets \L T=(\L S_t, \L Q_t)$
      \ENDFOR
      \ENDFOR
    \end{algorithmic}
\end{algorithm}

\subsection{Approaches to few-shot AED}
Few-shot learning does not prescribe a specific training procedure. We will discuss in the following how to tackle this problem under current setup from two aspects: (1). with traditional supervised methods, (2). with meta-learning approaches. 

\vspace{-0.1in}
\subsubsection{Supervised Baselines}
Before diving into meta-learning, it is important to explore solutions based purely on supervised learning. A natural approach to exploit $\L M_{train}$ would be to train a model $f$ over the whole training set $\L M_{train}$ by aggregating all training classes in $C_{train}$.
At test time, $f$ will be fine-tuned with the support set of new task $\L T$. Commonly, $f$ is prone to over-fitting at fine-tuning stage since $\L S$ only has few samples per event. A natural extension will be to fine tune only part of model. Formally if pre-trained model $f=[f_\theta, f_\phi]$, where $f_\theta$ is the feature extraction part and $f_\phi$ is classifier (commonly last linear layer), we would freeze $f_\theta$ and fine-tune only $f_\phi$ at test time. We refer the above two baselines as \textbf{FT-All} and \textbf{FT-Linear} respectively.

In addition to above two fine-tuning baselines, we also consider following supervised approach, which classifies query sample based on its distance to support samples in feature space.
Formally, given a new task $\L T=(\L S, \L Q)\in\L M_{test}$, $\forall \v x^{(q)}\in \L Q$, we define following distance as in equation \ref{eq:nn_base_dist}, which characterize how far $\v x^{(q)}$ is from being positive and negative.

\begin{equation}
  \label{eq:nn_base_dist}
  \begin{split}
    & d_k^t(\v x^{(q)};\theta,\L S)=\frac{\displaystyle\sum_{(\v x^{(s)}, \v y)\in\L S}dist(f_\theta(\v x^{(s)}), f_\theta(\v x^{(q)}))}{\sum_{(\v x^{(s)}, \v y)\in\L S}1_{y_k=t}} \\
    & t\in\{0,1\}, k\in\{1,..., K\}\\
    \end{split}
\end{equation}

$d_k^t(\v x^{(q)};\theta,\L S)$ is average distance of query sample $\v x^{(q)}$ to positive or negative samples of event $k$ in support set in the feature space. $dist(\cdot, \cdot)$ in equation \ref{eq:nn_base_dist} is a distance metric (e.g., L2, cosine), which can be tuned as hyper-parameter. For ease of evaluation, we further convert $d_k^t(\v x^{(q)};\theta,\L S)$ to probability $p_k^t(\v x^{(q)};\theta,\L S)$ using softmax function:

\begin{equation}
  \footnotesize
  \label{eq:nn_base_prob}
  p(y_k=t|\v x^{(q)};\theta,\L S)=\frac{\exp(-d_k^t(\v x^{(q)};\theta,\L S))}{\exp(-d_k^0(\v x^{(q)};\theta,\L S))+\exp(-d_k^1(\v x^{(q)};\theta,\L S))}
\end{equation}

Fine-tuning, which might potentially leads to over-fitting due to the small size of support set, is avoided in this method. Another motivation for this method is that embedding space learned from $\L M_{train}$ also provides meaningful representations for samples of unseen classes in $\L M_{test}$. We call this baseline {\bf NN}, which is short for nearest neighbor, as it classifies new data based on its neighboring data points. 

\vspace{-0.1in}
\subsubsection{Meta-learning Approaches} Meta-learning based models are trained end-to-end for the purpose of learning to build classifiers from few examples. At training time, we compute probability $p(\v y^{q}|\v x^{(q)}, f, \L S)$ for $\v x^{(q)}\in \L Q$. Since labels of query data ($\v y^{(q)}$) are known at training time, we can define loss function based on the ground-truth and prediction, thus being able to train model $f$ end-to-end via gradient descent.
Models vary by the manner in which the conditioning on support set is realized. Main approaches we experimented are described below. We will use same notations as above.

\textbf{Prototypical Networks} This approach is a simple integration of NN-baseline into the end-to-end meta-learning framework. 
Given task $\L T=(\L S, \L Q)$ we compute probability of query samples $\v x\in \L Q$ being positive or negative on event $k$, $p(y_k=t|\v x; \theta,\L S)$, as in equation \ref{eq:nn_base_dist} and \ref{eq:nn_base_prob}. At training time we use it to compute cross-entropy loss:

\begin{equation}
  \label{eq:proto_loss}
  L(\theta;\L T)=-\displaystyle\sum_{(\v x, \v y)\in\L Q}\displaystyle\sum_{k=1}^K w_k\log p(y_k|\v x;\theta,\L S)
\end{equation}

Model $f_\theta$ is updated via gradient descent. At test time, we follow equation \ref{eq:nn_base_dist} and \ref{eq:nn_base_prob} to compute $p(y_k=t|\v x; \theta,\L S)$ for any new data $\v x$. 

Above approach is an adaptation of prototypical network \cite{proto} from multiway-way classification to multi-label classification setting. Main difference between ours and \cite{proto} lies in the way of computing distance $d_k^t(\v x^{(q)};\theta,\L S)$. In \cite{proto}, the embeddings of data of class $k$ from support set are averaged as the prototype for that class and distance $d_k^t(\v x^{(q)};\theta,\L S)$ is between query sample and prototype. While here we compute distance of query sample $\v x^{(q)}$ to every support sample and then take the mean distance. In short, we average distance instead of averaging embedding. Such a modification is because negative samples of any event can come from much larger number of other classes. Mean embedding of negative samples might not be a good representation for negative class. 

\textbf{MetaOptNet} MetaOptNet is short for meta-learning via differentiable convex optimization, which is originally proposed in \cite{maml}. Similar to multi-prototypical network, MetaOptNet learns a classifier in feature space. However, a linear SVM classifier is learned instead of nearest neighbor:

\begin{equation}
  \label{eq:svm}
  \begin{split}
    & \min_{\v w}\min_{\pmb \xi}\frac{1}{2}\displaystyle\sum_{k=1}^K\displaystyle\sum_{t\in\{0,1\}}\|\v w_k^t\|^2+\lambda\displaystyle\sum_{k=1}^{K}\xi^k \\
    & \text{s.t. }\v w_{k}^{y_k}f_\theta(\v x)-\v w_{k}^t f_\theta(\v x) \geq 1-\delta_{y_k, t}-\xi^k \\
    & \forall k\in\{1,...,K\}, t\in\{0,1\},(\v x, \v y)\in\L S \\
  \end{split}
\end{equation}

We solve equation \ref{eq:svm} with a differentiable QP solver \cite{opt_net} so that $f_\theta$ can be learned in an end-to-end way. At test time, we solve equation \ref{eq:svm} to score new data $\v x^{(q)}$ and transform it to probability through equation \ref{eq:svm_prob}.

\begin{equation}
  \label{eq:svm_prob}
p(y_k=t|\v x^{(q)};\theta,\L S)=\frac{\exp(\v w_{k}^t f_\theta(\v x^{(q)}))}{\exp(\v w_{k}^1 f_\theta(\v x^{(q)}))+\exp(\v w_{k}^0 f_\theta(\v x^{(q)}))}
\end{equation}

MetaOptNet is explored here because SVM naturally fits binary classification setting. Additionally, it has achieved  state-of-the-art results in few-shot image classification benchmarks \cite{mini_imagenet,tiered_imagenet,cifar_fs,fc100}. 

\textbf{MAML} Model Agnostic Meta-Learning (MAML) \cite{meta_opt} is another popular meta-learning framework. Different from Prototypical network and MetaOptNet where classifier is trained on an embedding space, MAML learns initialization parameters $\theta_0$ and $\phi_0$ from meta training set $\L M_{train}$ such that the model can perform well on query set after a few steps of gradient descent. Support set $\L S$ are used to calculate loss used for gradient computation. Suppose model $f$ is initialized as $f_{\theta_0,\phi_0}$, let $\theta_N, \phi_N=\text{GD}(\theta_0,\phi_0;L, \L S,N)$ be the model parameters updated through $N$ steps of gradient descent where the loss function is $L$ (same as equation \ref{eq:sup_loss}) computed on support set $\L S$. We solve optimization problem defined as equation \ref{eq:maml}, which minimizes the cross-entropy loss of $f_{\theta_N,\phi_N}$ on query set $\L Q$.

\begin{equation}
  \label{eq:maml}
  \min_{\theta_0,\phi_0}L(\theta_N,\phi_N; \L Q)=\min_{\theta_0,\phi_0}L(\text{GD}(\theta_0,\phi_0;L, \L S,N);\L Q)
\end{equation}
Note only initial model parameters $\theta_0,\phi_0$ are updated throughout training process while $\theta_{1:N},\phi_{1:N}$ are just intermediate variables. Training in MAML is made possible by unrolling gradient descent steps. Given data $\v x^{(q)}\in\L Q$ at test time, we can compute the overall probability vector as $p(\v y|\v x^{(q)};\L S)=f_{\theta_N,\phi_N}(\v x^{(q)})$ where $\theta_N,\phi_N=\text{GD}(\theta_0,\phi_0;L, \L S,N)$.

\section{Experimental Setup}

We use Audioset \cite{audioset} for experiments. In total, there are 5.8k hours of audios from 527 sound classes in Audioset. The large number of sound events provides good test-bed for few-shot learning. The whole set of event classes of Audioset forms tree-like structure. We only select ``leaf'' nodes to ensure fine granularity of audio events. Besides, audios with annotation accuracy\footnote{According to https://research.google.com/audioset/dataset/index.html} less than $80\%$ are not selected. In total we obtained 19,841 10-second audio clips from 142 events. Following our setup, they are randomly split into meta training set with 99 events, meta validation set with 21 events and meta test set with 21 events. Table \ref{tab:example_event} shows event examples in each partition.

\begin{table}[h]
  \footnotesize
  \setlength{\tabcolsep}{2pt}
  \begin{tabular}{c||c|c|c}\hline
    & Train & Val & Test \\ \hline
    Events & \begin{tabular}{c}'Sneeze', 'Spray', \\ 'Electric guitar'\\'Choir','Cheering'\end{tabular} & \begin{tabular}{c}'Glockenspiel',\\'Toilet flush', 'Sink'\\ 'Vacumn cleaner'\end{tabular} & \begin{tabular}{c} 'Ukulele' \\ 'Clapping', 'Toot'\\,'Purr', 'Racing'\end{tabular} \\ \hline
  \end{tabular}
  \caption{\label{tab:example_event}Examples of events in train, validation and test set}
  \vspace{-0.1in}
\end{table}

For audio pre-processing, we compute log Mel-filterbank energy feature for each audio clip. It is calculated with window size of 25 ms and hop size of 10 ms. The number of mel coefficients is 64, which gives us log-mel spectrogram feature of size $1000\times 64$ for each audio clip. Global CMVN (Cepstral Mean and Variance Normalization) is applied on feature map before it is fed as input to the model.

We use convolutional neural network (CNN) with 4 blocks as our backbone model architecture. Each block is composed of $3\times 3$ convolution, ReLU, Batch Norm and $3\times 3$ max-pooling layers. We tuned CNN achitecture on meta validation set and did not find improvements from using deeper networks like ResNet \cite{res}. 

For few-shot setup, we experimented on 1-shot and 5-shot setting, as is commonly done in few-shot benchmarks \cite{mini_imagenet,tiered_imagenet,cifar_fs}. 
The number of ways is set to 5. The query set is composed of 15 positive samples per way. On the other hand, we sampled 10 and 50 negative data for support set respectively in 1-shot and 5-shot setting. The number of negative data is 150 for query set in both settings. Here the negative data are audios where none of 5 target events occur.  
We randomly sample 5000 training tasks, 200 validation tasks and 200 test tasks. For evaluation, we measure Area Under Curve (AUC) for Receiver Operating Characteristic (ROC) curve. AUC score averaged over tasks in test set is reported. Higher AUC is better.

For supervised baselines, we pre-train CNN model (same architecture as above) with all training data by doing 99-way detection. A small subset (10\%) is held out for purpose of hyper-parameter tuning. After pre-training, the last linear layer is replaced by a randomly initialized linear layer with output dimension being 5, which is equal to number of ways. Either the whole model (\textbf{FT-whole}) or the linear layer (\textbf{FT-Linear}) is fine-tuned for 20 epochs on the test task. For \textbf{NN} baseline, the distance metric is chosen among\{L2, cosine, dot\} and we find cosine distance works best.
For \textbf{MAML}, the model runs through 5 steps of gradient descent both at training and testing time. For prototypical network and MetaOptNet, number of detection ways is tuned to be 5 during training.

\section{Results}
\subsection{Main Results}
\begin{figure}[h]
   \centering
   \vspace{-0.15in}
   \includegraphics[width=\linewidth]{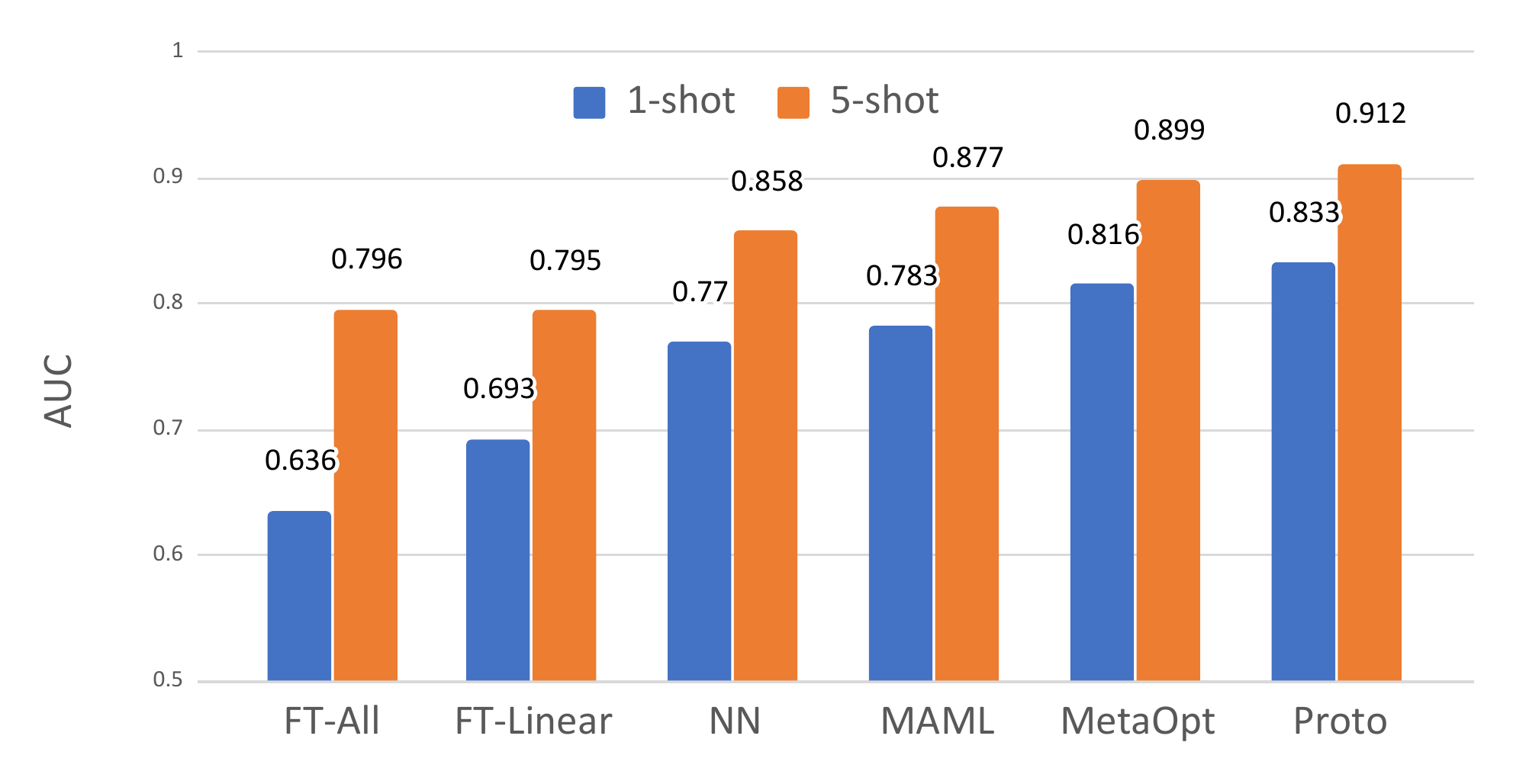}
   \caption{\label{fig:main} Mean AUC on test set of 3 supervised baselines and 3 meta-learning approaches. Higher is better}
   \vspace{-0.1in}
 \end{figure}

Figure \ref{fig:main} shows the performance of 6 methods on test set. Three meta-learning approaches outperform all supervised baselines on both 1-shot and 5-shot setting.
Meta-learning approaches are encouraged to generalize to unseen events as new task is sampled for training in each step.
Compared to supervised baselines, it is more effective on capturing the relationship between different events.

On the other hand, performance order of three supervised approaches are: FT-All$<$FT-Linear$<$NN, which implies less parameter tuning leads to better performance. In few-shot setting where labeled data are extremely scarce for one task, parameter fine-tuning often induces overfitting. As data amount increases we notice the gap between heavily fine-tuned approach and moderately tuned approach is reduced (FT-All vs. FT-Linear on 5-shot). Similar trend is shown on the side of meta-learning (MAML vs. Prototypical network/MetaOptNet).

 We also notice NN achieves comparable performance to meta-learning approaches despite its simplicity. This implies embedding function $f_\theta$ learned from detecting known events also equips audios with meaningful features on distinguishing unseen events because event classes are correlated with each other.
 For instance, ``orchestra'' and ``electric guitar'' both belong to music sound. A model trained on detecting ``orchestra'' sound captures characteristics of music, which can be used in detecting ``electric guitar''.  However, such model tends to overfit to over fine-grained features of known events. Model capacity on detecting distant events (e.g., those from non-music branch) will diminish. This issue can be mitigated by prototypical network, which also does nearest neighbor classification but learns embedding function $f_\theta$ through sampling from large pool of tasks, which helps generalization to new events.


 \subsection{Analysis}

 \textbf{Impact of domain mismatch}
 In default setup, we randomly split the events, under which the distribution of audio events in training and test set are the same. Here we created a setting where testing events come from a different distribution. For instance, all testing events belong to animal sounds while training events are household sounds.
 Such a setup is for testing robustness of meta-learning methods.
 More concretely, based on original setup we select one target domain (e.g., music), which is child of root node in Audioset ontology.
 All leaf nodes under target domain will be treated as new test set. Those events and their associated audios are removed from original training set. We re-run experiments for all 6 approaches under this setup.
 For purpose of comparison, we also evaluate same models with test set from which target events are removed. We experimented ``music'' and ``animal'' domains. Results are shown in table \ref{tab:ood}.

 \begin{table}[h]
  \setlength{\tabcolsep}{2pt}
   \centering
   \small
   \begin{tabular}{c||ccc|ccc}
      & \multicolumn{3}{|c|}{1-shot} & \multicolumn{3}{|c}{5-shot} \\
     AUC &  In-domain & Music & Animal & In-domain & Music & Animal \\ \hline
     FT-All	& 0.645 &	0.653 & 0.568 &	0.739 &	0.743 & 0.639 \\
     FT-Linear	& 0.72 &	0.716 & 0.612 &	0.798 &	0.785 & 0.670 \\
     NN	& 0.727 &	0.709 &	0.613 & 0.824 &	0.798 & 0.695 \\
     MAML &	0.715	& 0.651 & 0.599 &	0.826 &	0.780 & 0.666 \\
     MetaOpt	& 0.773& 0.699 & 0.640 &	0.856 &	0.81 & 0.740 \\
Proto	& \textbf{0.796} &	\textbf{0.712} & \textbf{0.644} &	\textbf{0.875} &	\textbf{0.824} & \textbf{0.749} \\ \hline
   \end{tabular}
   \caption{\label{tab:ood} Impact of domain discrepancy on different models. \textbf{Note}: ``In-domain'' numbers are different from those of figure \ref{fig:main} as both training and evaluation sets differ.}
\vspace{-0.2in}
 \end{table}
 
 Through the comparison of in-domain and target domain evaluation from table \ref{tab:ood}, all methods deteriorate due to domain mismatch. Results on ``music'' are slightly better than those on ``animal'', which is mainly because in Audioset music sounds are common and some patterns exist in other sounds as well such as bell ringing. On the other hand animal sounds such as ``dog growling'' rarely resemble other sounds. The gain of meta-learning approaches over supervised baselines are diminished due to domain mismatch. This implies the potential overfitting issue in meta-learning. Meta-learning models learn to utilize the correlation between classes. However, if all training classes come from one domain, the model tends to overfit to that particular domain and performance on new domain would drop.
 We notice that FT-linear achieves better results than MAML, which suggests learned features can still benefit detection in different domain. This can also be seen from overall high performance of feature-based approaches including NN, MetaOptNet and prototypical network. 

\textbf{Does pre-training help?} 
Compared to random guess, supervised baselines lead to much better results. Such a fact shows benefit of pre-training since all three baselines are built on the pre-trained detector.
Meta-learning models are trained on individual tasks where labeled data in each task are of small amount. We want to see whether pre-training with whole training set would also provide good starting point for meta-learning. We follow same pre-training steps as supervised baselines, which is pre-training a 99-way detector with whole training set. Model parameters except the last layer are used as initial parameters for each method. 
Results of pre-training vs. non-pretraining for three meta-learning approaches are shown in table \ref{tab:pre-train}.

\begin{table}[h]
  \centering
  \begin{tabular}{c||c|c|c} \hline
    AUC & MAML (+P) & MetaOpt (+P) & Proto (+P) \\ \hline
    1-shot & 0.783 (\textbf{0.788}) & 0.816 (\textbf{0.823}) & \textbf{0.833} (0.827) \\ \hline
    5-shot & \textbf{0.877} (0.841) & \textbf{0.899 (0.899)} & \textbf{0.912 (0.912)} \\ \hline
  \end{tabular}
  \caption{\label{tab:pre-train} Supervised pre-trained initialization vs. randomly initialization for meta-learning models (+P: with pre-training)}
  \vspace{-0.15in}
\end{table}

We do not find consistent gain from pre-training according to table \ref{tab:pre-train}. By learning with sub-tasks in training set, meta-learning models implicitly learns detector conditioned on support set despite of limited number of labeled data in each task. Initialization has little effect on the final model. Another possible reason is that our backbone CNN is of small size thus there is little overfitting in learning new task.

\section{Conclusion}

We formulated and studied few-shot acoustic event detection by comparing typical meta-learning and supervised approaches. Through experimentation, we find meta-learning approaches outperforms their supervised counterparts, which shows the effectiveness of its training setup on generalization to new audio events. Besides, we find methods based on feature learning (e.g., NN baseline, prototypical network) outperform others. Thus one line of our future work is to incorporate unlabeled audio data for representation learning in few-shot AED. Domain discrepancy has negative impact on performance of all methods we experimented. However, we can still achieve gains from some meta-learning methods. How to train an AED model more robust to domain mismatch under few-shot setting remains to be investigated. 



\bibliographystyle{IEEEbib}
\bibliography{strings,refs}

\end{document}